\pdfoutput=1

\documentclass[11pt]{article}
\usepackage{arabtex}
\usepackage[]{EMNLP2023}
\usepackage{lipsum}
\usepackage{times}
\usepackage{latexsym}

\usepackage[T1]{fontenc}

\usepackage[utf8]{inputenc}
\usepackage{xstring}
\usepackage{cleveref}
 \usepackage{arabtex}
\usepackage{utf8}
\usepackage{color}
\usepackage{arydshln}
\usepackage{adjustbox}
\usepackage{float}
\restylefloat{table}
\usepackage{array,booktabs,makecell}
\usepackage{multirow}
\usepackage{graphicx}
\usepackage{color, colortbl}
\usepackage{xstring}
\usepackage{arydshln}
\usepackage{microtype}
\usepackage{color,soul}

\usepackage{multirow}
\usepackage{graphicx}

\usepackage{listings}
\usepackage{color}

\definecolor{dkgreen}{rgb}{0,0.6,0}
\definecolor{gray}{rgb}{0.5,0.5,0.5}
\definecolor{mauve}{rgb}{0.58,0,0.82}

\usepackage{inconsolata}
\usepackage{arydshln}
\newcolumntype{H}{>{\setbox0=\hbox\bgroup}c<{\egroup}@{}}
\usepackage{cancel}
\usepackage{ulem,xpatch}

\usepackage{hyperref}

\definecolor{OCCcolor}{RGB}{42,96,160}

\definecolor{GPEcolor}{RGB}{57,108,29}
\def\GPE{\color{GPEcolor}}
\definecolor{ORGcolor}{RGB}{140,32,57}
\def\ORG{\color{ORGcolor}}
\def\b{\color{black}}
\def\o{\color{orange}}

\usepackage{enumitem}

\usepackage{tikz}
\def\checkmark{\tikz\fill[scale=0.4](0,.35) -- (.25,0) -- (1,.7) -- (.25,.15) -- cycle;} 
\usepackage[colorinlistoftodos,prependcaption,textsize=tiny]{todonotes}

\title{WojoodNER 2023: \\ The First Arabic Named Entity Recognition Shared Task}

\author{\normalsize Mustafa Jarrar$^{1}$ ~ Muhammad Abdul-Mageed$^{2,3}$ ~ Mohammed Khalilia$^{1}$ ~ \textbf{Bashar Talafha}$^{2}$\\
\normalsize \textbf{AbdelRahim Elmadany}$^{2}$ ~ \textbf{Nagham Hamad}$^{1}$~ \textbf{Alaa' Omar}$^{1}$ \\
\normalsize $^{1}$Birzeit University, Palestine\\
  \normalsize $^{2}$Deep Learning \& Natural Language Processing Group,
  The University of British Columbia\\
  \normalsize  $^{3}$Department of Natural Language Processing \& Department of Machine Learning, MBZUAI\\ %
  \texttt{\normalsize mjarrar@birzeit.edu ~ ~ muhammad.mageed@ubc.ca}
}
  
\begin{document}
\setcode{utf8}
\setarab 
\maketitle

\begin{abstract}

We present WojoodNER-$2023$, the first Arabic Named Entity Recognition (NER) Shared Task.  The primary focus of WojoodNER $2023$ is on Arabic NER, offering novel NER datasets (i.e., Wojood) and the definition of subtasks designed to facilitate meaningful comparisons between different NER approaches. WojoodNER-$2023$ encompassed two Subtasks: FlatNER and NestedNER. A total of $45$ unique teams registered for this shared task, with $11$ of them actively participating in the test phase. Specifically, $11$ teams participated in FlatNER, while $8$ teams tackled NestedNER. The winning teams achieved $F_1$ scores of $91.96$ and $93.73 $ in FlatNER and NestedNER, respectively.

\end{abstract}

\section{Introduction}

NER is a fundamental task in Natural Language Processing (NLP), especially in information extraction and language understanding \cite{JMHK23}. The objective of NER is to identify and classify named entities in a given text into predefined categories, such as ``\textit{person}'', ``\textit{location}'', ``\textit{organization}'', ``\textit{event}'', and ``\textit{occupation}''. NER is also a critical task for many NLP applications, such as question-answering systems \cite{shaheen2014arabic}, knowledge graphs \cite{james1992knowledge}, and semantic search \cite{guha2003semantic}, interoperability \cite{JDF11}  among
others. Named entities can either be flat or nested. 
For instance, in the sentence ``\texttt{Cairo  Bank announces its profit in $2023$}'', there are two flat entities: ``\texttt{\ORG Cairo Bank\b}'' is tagged as {\ORG \texttt{ORG}\b} (i.e., organization) and ``\texttt{\o$2023$\b}'' as {\o \texttt{DATE}\b}. 
In nested NER, entity mentions contained inside other entity mentions are also considered named entities. In this case,  ``\texttt{\GPE Cairo\b}'', is tagged as { \GPE \texttt{GPE}\b} (i.e., geopolitical entity). Section \ref{sec:task_description} illustrates more examples. As will be discussed in Section~\ref{sec:literture}, research in Arabic NER  is currently limited, particularly in the context of nested entities. This limitation is not exclusive to Modern Standard Arabic (MSA) but extends to various Arabic dialects across diverse domains and NER subtypes. The majority of existing research on Arabic NER primarily emphasizes flat entities to cover a limited set of entity types, mainly ``person'', ``organization'', and ``location''.

\begin{figure}[t]
  \begin{center}
  \includegraphics[width=0.95\linewidth]{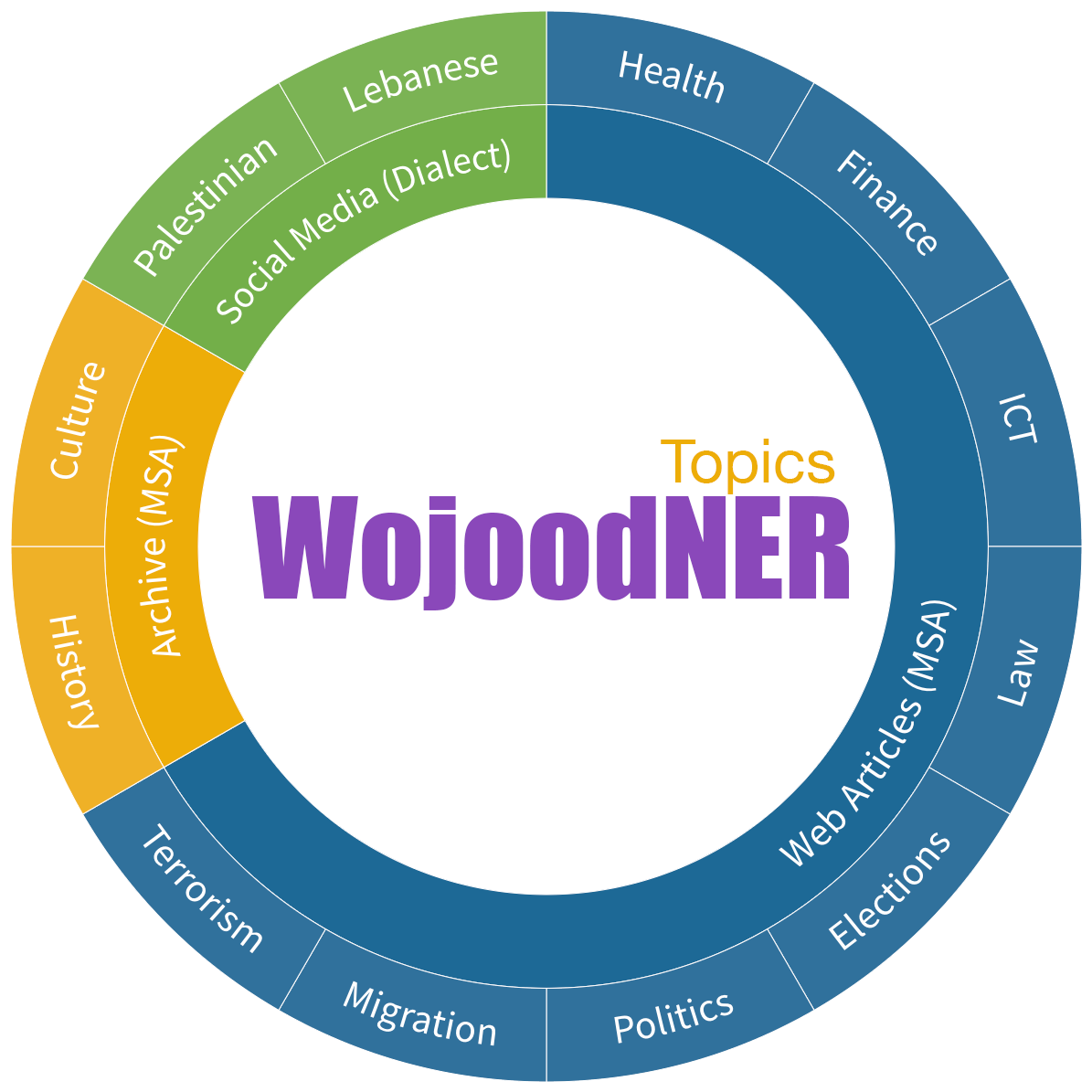}
  \end{center}
\caption{Topics in the Wojood NER corpus.}
\label{fig:examples}
\end{figure}


In this paper, we provide an overview of the WojoodNER-$2023$ Shared Task\footnote{{\scriptsize SharedTask Call: }\url{https://dlnlp.ai/st/wojood/}}, which represents a significant step forward in advancing NER research in the Arabic language. The shared task encompasses subtask1 (FlatNER) and subtask2 (NestedNER). For this competition, we grant participants access to the Wojood corpus~\cite{JKG22}\footnote{{\scriptsize Wojood Corpus: } \url{https://sina.birzeit.edu/wojood/}}, a substantial and diverse Arabic NER dataset known as Wojood. As shown in Figure \ref{fig:examples}, Wojood is particularly notable for its scale, containing approximately $550$K tokens. About $12$\% of the corpus was collected from social media in \textit{Palestinian} and \textit{Lebanese} dialects Curras and Baladi corpora \cite{EJHZ22}. The remaining $\sim88$\% is in MSA, covering multiple domains, including \textit{health}, \textit{finance}, \textit{politics}, \textit{ICT}, \textit{terrorism}, \textit{migration}, \textit{history and culture}, and \textit{law and elections}, making it a rich resource for various research purposes. Wojood was annotated manually using $21$ entity types, offering a rich Arabic NER corpus.

The primary objective of this shared task is to encourage participants to explore different NER methodologies. Teams were invited to experiment with various approaches, ranging from classical machine learning to advanced deep learning and transformer-based techniques, among others. The shared task generated a remarkably diverse array of submissions. A total of $45$ teams registered to participate in the shared task. Among these, $11$ teams successfully submitted their models for evaluation on the blind test set during the final phase of the competition. As a result, we received $11$ papers that provide detailed insights into the results achieved by these teams for either one or both of the subtasks.


The rest of the paper is organized as follows: Section~\ref{sec:literture} provides a brief overview of Arabic NER. We describe the two subtasks and WojoodNER-2023 restrictions in Section~\ref{sec:task_description}. Section~\ref{sec:eval} introduces shared task datasets and evaluation setup. We present participating teams and shared task results and provide
a high-level description of submitted systems in Section~\ref{sec:teams_results}. We conclude in Section~\ref{sec:conc}.

\section{Literature Review} \label{sec:literture}

NER has been a long-standing research area, with significant advances made in recent years.  
As will be discussed in this section, early NER approaches focused on identifying and classifying flat named entities, and recent research has focused on nested NER. In this section, our primary focus is exclusively on Arabic NER research, encompassing corpora, methodologies and shared tasks.

\paragraph{Corpora.} Most of the available Arabic NER corpora are annotated as flat NER. \texttt{ANERCorp}~\cite{benajiba2007anersys}, sourced from the news domain (MSA text), comprises $\sim150$k tokens. Its main emphasis is directed towards four distinct entity types. \texttt{CANERCorpus}~\cite{salah2018building} is dedicated to Classical Arabic (CA) and encompasses a dataset of $258$K tokens. This corpus is annotated for a total of $14$ entity types, all of which pertain to religious entities. \texttt{ACE2005}~\cite{ACE2005} is a multilingual corpus that incorporates Arabic text encompassing \textit{five} distinct types of entities. \texttt{Ontonotes5}~\cite{ontonote} dataset consists of approximately $300$K tokens, meticulously annotated with $18$ distinct entity types. Nevertheless, these corpora were collected a long time ago and mainly cover the media and politics domains; hence, may not be representative of the current state of Arabic language use. This is especially the case since language models are known to be sensitive to temporal and domain shifts. Recently,~\newcite{JKG22} proposed \texttt{Wojood}, the largest Arabic NER corpus. It is distinctive for its support of both flat and nested entity annotations, making it a crucial resource utilized in this shared task. It comprises roughly $550$K tokens encompassing a diverse range of $21$ unique entity types, spanning both MSA and two dialectal Arabic forms (the Palestinian Curras2 and Lebanese Baladi corpora \cite{EJHZ22}.

\paragraph{Methodologies.} Various studies explore Arabic NER by employing various approaches, with some researchers focusing on rule-based \cite{shaalan2007person,jaber2017morphology} and machine learning~\cite{settles2004biomedical,abdul-hamid-darwish-2010-simplified,zirikly-diab-2014-named,dahan2015first,DH21} strategies. Recent researches embrace deep learning methodologies including character and word embeddings with Long-Short Term Memory (LSTM) networks~\cite{ali2018bidirectional}, BiLSTM followed by Conditional Random Field (CRF) models~\cite{el2019arabic, khalifa2019character},  Deep Neural Networks (DNN)~\cite{gridach2018deep}, and pretrained Language Models (LM)~\cite{JKG22, LJKOA23}. \newcite{wang2022nested} proposed a survey that extensively explores different approaches to nested entity recognition, encompassing rule-based, layered-based, region-based, hypergraph-based, and transition-based methodologies. \citet{fei2020dispatched} proposed a multitask learning approach for nested NER that employs a dispatched attention model. \citet{ouchi2020instance} proposed an approach for nested NER that involves enumerating all region representations from the contextual encoding sequence and then assigning a category label to each of them.

\paragraph{Shared tasks.} While there are multiple shared tasks for NER in various languages and domains, such as the \texttt{MultiCoNER} for multilingual complex NER ~\cite{malmasi-etal-2022-semeval}, the \texttt{HIPE-2022} for NER and linking in multilingual historical documents \cite{HIPE-2022}, the \texttt{RuNNE-2022} for nested NER in Russian \cite{RuNNE-2022}, and the \texttt{NLPCC2022} for extracting entities in the material science domain \cite{NLPCC2022}.
To the best of our knowledge, there has been no dedicated shared task for Arabic NER. Therefore, we initiate this shared task with the aim of being the inaugural event in this specific domain.

\section{Task Description}\label{sec:task_description}



To the best of our knowledge, WojoodNER-$2023$ is recognized as the inaugural shared task in Arabic NER. In this competition, we present two distinct subtasks—one for ``\textbf{FlatNER}'' and the other for ``\textbf{NestedNER}''. These subtasks are of paramount importance in addressing the challenges inherent in Arabic NER processing. We now describe each subtask in detail.

\subsection{Subtask1 -- FlatNER}
\begin{figure}[h!]  
    \centering       
    \includegraphics[width=0.45\textwidth]{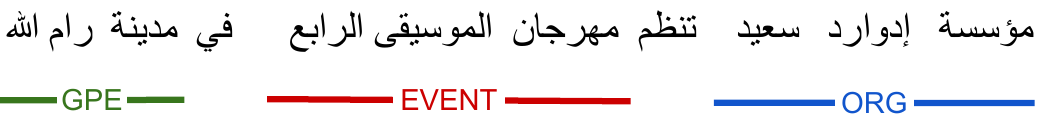}
   \caption{Flat NER example}
\label{fig:flat_ner_example}
\end{figure}

In FlatNER, each token in the data is labeled with only one tag. The participants in this subtask are expected to develop models to classify each token as a multi-class classification problem. An example of the FlatNER data is shown in Figure \ref{fig:flat_ner_example}. The Wojood annotation guidelines were designed for nested entities only, therefore, the flat entities were derived from the nested entities by taking the top-level entity mentions (i.e., topmost tags).


\subsection{Subtask2 -- NestedNER}
\begin{figure}[h!]  
    \centering
    \includegraphics[width=0.48\textwidth]{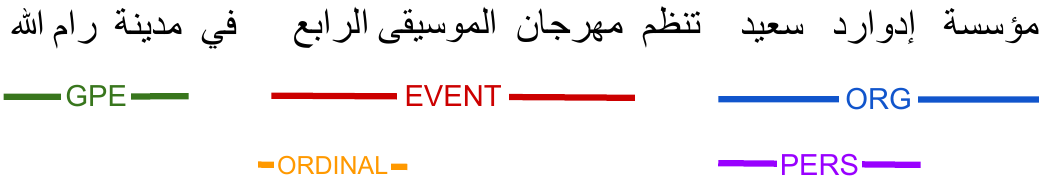}
   \caption{Nested NER example}
\label{fig:nested_ner_example}
\end{figure}

In the NestedNER subtask, each token can have one or more tags. In this data, we will find entity mentions inside other entity mentions as demonstrated in Figure \ref{fig:nested_ner_example}. For instance, the phrase {\small``\<مؤسسة إدوارد سعيد>''} is annotated as  \texttt{ORG}, which is the same as the flat annotation in Figure \ref{fig:flat_ner_example}. However, in nested NER, it contains another entity mention {\small``\< إدوارد سعيد>''} tagged with \texttt{PERS}.


\subsection{Restrictions}
This section outlines the stipulations and directives that govern participants' engagement in the WojoodNER $2023$ Shared Task. These regulatory directives and guidelines establish an equitable competitive environment for all participants, ensuring transparency and impartiality throughout the duration of the WojoodNER $2023$ Shared Task. They also ensure the credibility of the task's assessment procedure, which was published on the shared task official website frequently asked question page.

\paragraph{External data.} Participants are strictly prohibited from using external data from previously labeled datasets or employing taggers that have been previously trained to predict named entities. The use of any resources with prior knowledge related to NER is not allowed.

\paragraph{Data format constraints.} The submission to the task consists of one file containing the model prediction in CoNLL format. The CoNLL format should include multiple columns space-separated. The first column is reserved for the tokens, while all subsequent columns are used for the tags. In the case of nested NER, the tag columns have a predefined order, which we specified on the shared task webpage\footnote{\url{https://dlnlp.ai/st/wojood/}}. The IOB2 \citep{sang1999representing} scheme is used for the submission, which is the same format used in the Wojood dataset. Finally, text segments are separated by a blank line.

\paragraph{Pretrained models.} The participants are allowed to utilize pretrained transformer models such as ``\textit{BERT}'' \cite{devlin2018bert} and word representations like ``\textit{Word2Vec}'' ~\cite{church2017word2vec} and ``\textit{ELMo}'' ~\cite{peters-etal-2018-deep} for the purpose of transfer learning. It is worth noting that our baseline model is based on BERT.

\paragraph{Linguistic features.} When considering the incorporation of linguistic features to enhance the dataset, participants are permitted to include part-of-speech tagging and syntactic layers within their code.
\section{Shared Task Datasets and Evaluation} \label{sec:eval}
This section presents the dataset, evaluation metrics, and the submission process.

\begin{table*}[h!]
\centering
\resizebox{\textwidth}{!}{%
\begin{tabular}{l|l|rrrrr|rrrr}
\toprule
\multirow{2}{*}{\textbf{Entity Name}} & \multirow{2}{*}{\textbf{NER Tag}} & \multicolumn{4}{c}{\textbf{FlatNER}} & \multirow{2}{*}{} & \multicolumn{4}{c}{\textbf{NestedNER}} \\ \cmidrule{3-6}\cmidrule{8-11}
 & & \textbf{{TRAIN}} & \textbf{{DEV}} & \textbf{{TEST}} & \textbf{{Total}} &  & \textbf{{TRAIN}} & \textbf{{DEV}} & \textbf{{TEST}} & \textbf{{Total}} \\ \midrule

Person & \texttt{PERS}  & $4,496$ & $650$ & $1,409$ & $ \bf 6,555$ & & $4,994$ & $730$ & $1,562$ & $\bf 7,286$ \\
Group of people & \texttt{NORP} & $3,505$ &  $488$ & $948$ & $ \bf 4,941$& & $3747$ & $520$ & $1006$ & $\bf 5273$  \\
Occupation & \texttt{OCC} & $3,774$ & $544$ & $1,058$ & $\bf 5,376$ & &$3,887$ & $551$ & $1,95$ & $\bf 5,533$  \\
Organization & \texttt{ORG} & $10,731$ & $1,566$ & $3,047$ & $\bf 15,344$ && $13,174$ & $1,869$ & $3,738$ & $\bf 18,781$  \\
GeoPolitical Entity & \texttt{GPE} & $8,133$ & $1,132$ & $2,281$ & $\bf 11,546$ & & $15,300$ & $2,163$ & $4,315$ & $\bf 21,778$  \\
Geographical location& \texttt{LOC} & $510$ & $63$ & $168$ &$\bf 741$  && $619$ & $76$ & $204$ & $\bf 899$ \\
Facility (e.g., landmarks) & \texttt{FAC} & $689$ & 	$85$ &	$165$	&$\bf 939$	&&	$880$&	$111$&	$224$&	$\bf 1,215$ \\
Product & \texttt{PRODUCT}  & $ 36 $ & $5$ & $13$ & $\bf 54 $ &  & $	36 $ & $ 5 $ & $ 14 $ & $\bf 55$  \\
Event & \texttt{EVENT}  & $ 1,863	 $ & $ 253 $ & $	556 $ & $\bf	2,672 $ &  & $	1,934 $ & $	267 $ & $	577 $ & $\bf	2,778$\\
Date & \texttt{DATE}  & $ 10,667 $ & $	1,567 $ & $	3,091 $ & $\bf	15,325	$ &  & $11,290 $ & $	1,656 $ & $	3,288 $ & $\bf 	1,6234$\\
Time & \texttt{TIME}  & $ 286 $ & $	55 $ & $ 	84 $ & $\bf	425	$ &  & $288	 $ & $ 55$ & $ 	84$ & $\bf	427$\\
Language & \texttt{LANGUAGE}  & $ 131 $ & $	15 $ & $51 $ & $\bf	197$ &  & $		132 $ & $	15 $ & $	51 $ & $\bf	198$\\
Website & \texttt{WEBSITE}  & $ 434 $ & $	45 $ & $128 $ & $\bf	607	$ &  & $434 $ & $	45 $ & $	128 $ & $\bf	607$\\
Law & \texttt{LAW}  & $ 374 $ & $	44 $ & $	78 $ & $\bf	496 $ &  & $	374 $ & $	44 $ & $	78 $ & $\bf	496$\\
Cardinal & \texttt{CARDINAL}  & $ 1,245 $ & $	182 $ & $	360	$ & $\bf 1,787	$ &  & $1,263 $ & $	183 $ & $	363 $ & $\bf	1,809$\\
Ordinal & \texttt{ORDINAL}  & $ 2,805 $ & $	410 $ & $	858 $ & $\bf	4,073	$ &  & $3,488 $ & $	504 $ & $	1,070 $ & $\bf	5,062$   \\
Percent & \texttt{PERCENT}  & $ 105 $ & $	13 $ & $	19 $ & $\bf	137	$ &  & $105 $ & $	13 $ & $	19 $ & $\bf	137$ \\
Quantity & \texttt{QUANTITY}  & $ 44 $ & $	3 $ & $	7 $ & $\bf	54	 $ &  & $46 $ & $	3 $ & $	8 $ & $\bf	57 $ \\
Unit & \texttt{UNIT}  & $ 7 $ & $	0 $ & $	2 $ & $\bf	9 $ & $		$ & $48 $ & $	3 $ & $	9	 $ & $\bf 60 $ \\
Money & \texttt{MONEY}  & $ 171 $ & $	20 $ & $	36 $ & $\bf	227	$ &  & $	171 $ & $	20 $ & $	36 $ & $\bf	227 $\\
Currency & \texttt{CURR}  & $ 19 $ & $	1 $ & $	5 $ & $\bf 25 $ & $		$ & $179 $ & $	21 $ & $	41 $ & $\bf	241 $ \\
\midrule
& \textbf{{Total}}&  $\bf 50,025$ & $\bf 7,141$ & $\bf 14,364$ & $\bf 71,530$ & & $\bf 62,389$ & $\bf 8,854$ & $\bf 17,910$ & $\bf 89,153$ \\
\bottomrule
\end{tabular}%
}
\caption{ Distribution of NER tags in WojoodNER-$2023$ Subtask1 (i.e., FlatNER)  and Subtask2 (i.e., NestedNER) across the training (i.e., TRAIN) , development (i.e., DEV), and test (i.e., TEST)  splits for the WojoodNER-$2023$. 
}
\label{tab:datasets}
\end{table*}


\paragraph{Datasets.} WojoodNER-$2023$ shared task employs the Wojood corpus as its primary dataset  \citep{JKG22}. The Wojood corpus encompasses approximately $550$K tokens, spanning both MSA and two Arabic dialects, annotated using $21$ entity types. Wojood annotation guidelines are optimized for nested Arabic NER annotations. However, for the purposes of the shared task, we generate a flat NER dataset by reducing the nested NER annotation to the top level only as demonstrated in Figure \ref{fig:flat_ner_example} and \ref{fig:nested_ner_example}. For both subtasks, we split the data $70$/$10$/$20$ for training, development, and test dataset respectively at the domain level.  This split ensures similar data distribution across the three datasets. Table~\ref{tab:datasets} present the statistics and characteristics
of WojoodNER-$2023$’s subtask1 and subtask2 training, development, and test datasets. 




\begin{table*}[!ht]
\centering
\resizebox{0.9\textwidth}{!}{%
\begin{tabular}{lll}
\toprule
\textbf{Team} & \textbf{Affiliation} & \textbf{Task} \\
\midrule
Alex-U $2023$ NLP  \cite{Hussein-etal-2023-AraBINDER}& Alexandria University & $1$,$2$\\
AlexU-AIC \cite{elkordi-etal-2023-sequence-mrc}& Alexandria University & $1$,$2$\\
AlphaBrains  \cite{ehsanwojoodner2023}& University of Gujrat, Pakistan & $1$,$2$ \\
ARATAL 
&IPSA&1 \\
El-Kawaref  \cite{El-Kawaref}& German University in Cairo& $1$ \\
ELYADATA \cite{ElyadataSubmission} & ELYADATA & $1$,$2$ \\
Fraunhofer IAIS  
& Fraunhofer IAIS &$1$ \\
LIPN  \cite{lipn-at-wojoodner-shared-task}& LIPN, Université Paris $13$ & $1$,$2$ \\
Lotus  \cite{lotus2023}& MBZUAI & $1$,$2$ \\
R00  
& Jordan University of Science and Technology & $1$,$2$ \\
Think NER  & Ulm University & $1$,$2$ \\
UM6P \& UL \cite{el-mahdaouy-etal-2023-um6p}& Mohammed VI Polytechnic University & $1$,$2$ \\
\bottomrule
\end{tabular}%
}
\caption{List of teams that participated in either one or both subtasks. Teams with accepted papers are cited. } 
\label{tab:teams}
\end{table*}
\paragraph{Evaluation metrics.} The official evaluation metric for subtask1 and subtask2 is the macro-averaged \texttt{F\textsubscript{1}} score. In addition to this metric, we also report system performance in terms of  \texttt{Precision}, \texttt{Recall}, and \texttt{Accuracy} for submissions to both subtasks.

\paragraph{Submission roles.} We allowed participant teams to submit up to \textit{four} runs for each test set, for both subtasks. In each one, we strictly retain only the submission with the highest score from each participating team. Although the official results were solely derived from the blind test set. To streamline the evaluation of participant systems, we have set up two separate CodaLab~\cite{codalab_competitions_JMLR} competitions for scoring each subtask.\footnote{The different CodaLab competitions are available at the following links: \href{https://codalab.lisn.upsaclay.fr/competitions/11740}{\texttt{subtask-1}} and \href{https://codalab.lisn.upsaclay.fr/competitions/11750}{\texttt{subtask-2}}.} We are keeping the CodaLab~\cite{codalab_competitions_JMLR} for each subtask active even after the official competition has concluded. This is aimed at facilitating researchers who wish to continue training models and evaluating systems with the shared task's blind test sets. As a result, we will not disclose the labels for the test sets in any of the subtasks.
\section{Shared Task Teams \& Results}\label{sec:teams_results}

\subsection{Participating Teams}

In total, we received $45$ unique team registrations. At the testing phase,  a total of $57$ valid entries were submitted by $12$ unique teams. We received $35$ submissions for FlatNER from \textit{eleven} teams and $22$ submissions for NestedNER from \textit{eight} teams. Table~\ref{tab:teams} lists the teams, their affiliation, and the tasks they participated in (Subtask1 -- FlatNER and Subtask2 -- NestedNER). From $12$ teams we received $11$ description papers from which we accepted $8$ for publication and $3$ were rejected (for quality or not adhering to the shared task guidelines).  


\subsection{Baselines}

For both subtasks, we fine-tune the AraBERT\textsubscript{v2} \cite{antoun2020arabert} and ARBERT\textsubscript{v2} \cite{abdulmageed2021arbert} pre-trained models using the training data that is specific to each subtask for $20$ epochs and employed a learning rate of $1e-5$, along with a batch size of $16$. To ensure model optimization, we incorporate early stopping with a patience setting of $5$. After each epoch, we evaluated the model's performance and selected the best-performing checkpoints based on their performance on the respective development set. Subsequently, we present the performance metrics of the best-performing model on the test datasets.

\begin{table}[!ht]
\centering
\resizebox{0.45\textwidth}{!}{%
\begin{tabular}{@{}clrrr@{}}
\toprule
\textbf{Rank} & \textbf{Team}   & \textbf{F1} & \textbf{Pre.} & \textbf{Rec.} \\ \midrule
1 &  LIPN            & $\bf 91.96$ & $92.56$ & $91.36$  \\
2 &  El-Kawaref      & $91.95$ & $91.43$ & $92.48$ \\
3 &  ELYADATA        & $91.92$ & $91.88$ & $91.96$  \\
4&  Alex-U 2023 NLP & $91.80$ & $91.61$ & $92.00$   \\
5 & Think NER  & $91.25$ & $90.76$ & $91.73$ \\
6 & ARATAL & $91.13$ & $90.49$ & $91.77$ \\
7 &  UM6P \& UL      & $91.13$ & $90.70$ & $91.57$  \\
8 &  AlexU-AIC       & $91.13$ & $91.33$ & $90.92$  \\
\cdashline{1-5}
& Baseline-I (ARBERT\textsubscript{v2} ) & $89.20$ & $88.32$ &$90.09$ \\
& Baseline-II (AraBERT\textsubscript{v2})  & $87.33$ & $86.00$ &$88.00$   \\\cdashline{1-5}

9&  AlphaBrains     & $87.15$ & $87.45$ & $87.58$ \\
10&  Lotus           & $83.39$ & $80.90$ & $86.04$ \\
11& R00 & $76.99$ & $76.67$ & $77.31$\\
12& Fraunhofer IAIS & $64.45$ & $65.53$ & $63.40$\\
\bottomrule
\end{tabular}%
}
\caption{Results of Subtask1 -- FlatNER. 
}
\label{tab:results1}
\end{table}

\subsection{Results}

\begin{table}[!ht]
\centering
\resizebox{0.45\textwidth}{!}{%
\begin{tabular}{clccc}
\toprule
\textbf{Rank} & \textbf{Team}   & \textbf{F1} & \textbf{Pre.} & \textbf{Rec.} \\ \midrule
1&  Elyadata        & $93.73$ & $ 93.99$ & $93.48$  \\
2&  UM6P \& UL      & $93.03$ & $92.46$ & $93.61$ \\
3&  AlexU-AIC       & $92.61$ & $92.10$ & $93.13$ \\
4&  LIPN            & $92.45$ & $92.31$ & $92.59$   \\
\cdashline{1-5}
& Baseline-I (ArBERT\textsubscript{v2})    & $91.68$ & $91.01$ & $92.35$    \\\cdashline{1-5}
5 & Think NER & $91.4$ & $90.03$ & $92.82$ \\ \cdashline{1-5}
& Baseline-II (AraBERT\textsubscript{v2} )    & $91.06$ & $90.74$ & $91.38$    \\\cdashline{1-5}
6&  Alex-U 2023 NLP & $90.01$ & $89.39$ & $90.63$   \\
7&  AlphaBrains     & $88.84$ & $88.45$ & $89.23$ \\
8&  Lotus           & $76.02$ & $82.19$ & $70.72$   \\
\bottomrule
\end{tabular}%
}
\caption{Results of Subtask2 -- NestedNER. }
\label{tab:results2}
\end{table}
\begin{table*}[ht!]
\resizebox{0.9\textwidth}{!}{%
\begin{tabular}{lHcccccccccccc}
\toprule

 \multirow{2}{*}{\textbf{Team Name}} & \multirow{2}{*}{\rotatebox[origin=c]{70}{\textbf{\# submissions}}}  & \multicolumn{1}{l}{} & \multirow{2}{*}{\rotatebox[origin=c]{70}{\textbf{Preprocessing}}} & \multicolumn{3}{c}{\textbf{Features}} & \multicolumn{7}{c}{\textbf{Techniques}} \\ \cmidrule(l){5-7} \cmidrule(l){8-14}
 & & \multicolumn{1}{l}{\multirow{-2}{*}{$\bf F_1$}} &  & \multicolumn{1}{c}{\rotatebox[origin=c]{70}{\textbf{TF-IDF}}} & \multicolumn{1}{c}{\rotatebox[origin=c]{70}{\textbf{Word Embeds}}} & \multicolumn{1}{c}{\rotatebox[origin=c]{70}{\textbf{Resampling}}} & \multicolumn{1}{c}{\rotatebox[origin=c]{70}{\textbf{Neural Nets}}} & \multicolumn{1}{c}{\rotatebox[origin=c]{70}{\textbf{Contrast. L}}} & \multicolumn{1}{c}{\rotatebox[origin=c]{70}{\textbf{Ensemble}}} & \multicolumn{1}{c}{\rotatebox[origin=c]{70}{\textbf{Adapter}}} & \multicolumn{1}{c}{\rotatebox[origin=c]{70}{\textbf{Multitask}}} & \multicolumn{1}{c}{\rotatebox[origin=c]{70}{\textbf{PLM}}} & \multicolumn{1}{c}{\rotatebox[origin=c]{70}{\textbf{Hie. Cls}}} \\ \midrule
\multicolumn{13}{c}{\textbf{FlatNER}}  &  \\ \midrule 
 
LIPN & -- & { 91.96} &  &  &  &  & \checkmark &  & \checkmark &  &  & \checkmark &  \\
{El-Kawaref} &  &  91.95 & \checkmark &  &  &  &  &  &  &  & & \checkmark  &  \\
 
Elyadata &  --& { 91.92} & { \checkmark} &  &  & \checkmark & \checkmark &  &  &  &  & \checkmark &  \\
 
Alex-U 2023 NLP &--  & { 91.80} &  &  &  &  &  & \checkmark &  &  &  & \textbf{\checkmark} &  \\
   ThinkNER &  -- & { 91.25} &  &  &  &  &  &  &  &  &  &  &  \\
UM6P \& UL & -- & { 91.13} &  &  &  &  &  &  &  &  & \checkmark & \checkmark &  \\

 AlexU-AIC &  & { 91.13} & \checkmark &  &  &  &  &  &  &  &  & \checkmark & \checkmark \\
ARATAL & -- & { 91.13} &  &  &  &  &  &  & \checkmark &  &  & \checkmark &  \\
AlphaBrains &  --& { 87.51} &  &  & \checkmark &  & \checkmark &  &  &  & \checkmark &  &  \\
 
Lotus & -- & { 83.39} & \checkmark & \checkmark &  &  &  &  &  & \checkmark &  & \checkmark &  \\
 
Fraunhofer IAIS & -- & { 64.45} &  &  &  &  &  &  &  &  &  & \checkmark &  \\
 
\midrule
\multicolumn{13}{c}{\textbf{NestedNER}} &  \\\midrule
 
Elyadata &  --& { 93.73} & { \checkmark} &  &  & \checkmark & \checkmark &  &  &  &  & \checkmark &  \\
 
UM6P \& UL & -- & { 93.03} &  &  &  &  &  &  &  &  & \checkmark & \checkmark &  \\

AlexU-AIC & -- & { 92.61} & \checkmark &  &  &  &  &  &  &  &  & \checkmark & \checkmark \\
 LIPN & -- & { 92.45} &  &  &  &  & \checkmark &  & \checkmark &  &  & \checkmark &  \\
 ThinkNER &  -- & { 91.40} &  &  &  &  &  &  &  &  &  &  &  \\
 Alex-U 2023 NLP &  -- & { 76.02} &  &  &  &  &  & \checkmark &  &  &  & \textbf{\checkmark} &  \\
AlphaBrains & -- & { 88.84} &  &  & \checkmark &  & \checkmark &  &  &  & \checkmark &  &  \\
 
Lotus& -- & { 76.02} & \checkmark & \checkmark &  &  &  &  &  & \checkmark &  & \checkmark &  \\
 
 \bottomrule
 
\end{tabular}%
}
\caption{Summary of approaches used by participating teams in subtask1 (i.e., FlatNER) and subtask2 (i.e., NestedNER). Teams are sorted by their performance on the official metric, Macro-$F_1$ score. The term ``Neural Nets" refers to any model based on neural networks (e.g., FFNN, RNN, CNN, and Transformer) trained from scratch. PLM refers to neural networks pretrained with unlabeled data such as ARBERT\textsubscript{v2}. (Hie. Cls, hierarchical classification approach); (Contrast. L, contrastive learning).}
\label{tab:teams_work}
\end{table*}
Table~\ref{tab:results1} and Table~\ref{tab:results2} present the leaderboards of Subtask1 -- FlatNER and Subtask2 -- NestedNER, respectively, sorted by macro-$F_1$ in descending order. The macro-$F_1$ score for each team represents the highest score among the four allowed submissions for each task. 

For FlatNER, \texttt{LIPN} team~\cite{lipn-at-wojoodner-shared-task} achieved the highest $F_1$ score of $91.96$, while \texttt{El-Kawaref}~\cite{El-Kawaref} came in second place with $91.95$ and \texttt{Elyadata} in third place with $91.92$. Notably, on FlatNER, \textit{eight} teams surpass our two baselines performance, as seen in Table \ref{tab:results1}. Moreover, the winning team (i.e, \texttt{LIPN}~\cite{lipn-at-wojoodner-shared-task}) outperforms the Baseline-I by $2.76$\%. \textit{Three} teams underperform Baseline-I and Baseline-II. However, the gap between the baseline-I and the worst-performing model is about $24.75$\%. We also notice that the difference in the $F_1$ score among the top \textit{eight} teams is marginal ($\sigma=0.41$).

We also analyzed the performance at the entity-type level in FlatNER and we noticed that certain entity types are more challenging to learn by all submitted models, including the baseline. The main reason for their low performance is the rarity of those entities in the dataset, with frequency reaching as low as $9$ for \texttt{UNIT} and $54$ for both \texttt{PRODUCT} and \texttt{QUANTITY}. The highest $F_1$ for \texttt{PRODUCT} is $61.54$ \cite{Hussein-etal-2023-AraBINDER}, for \texttt{QUANTITY} $50.00$ \cite{El-Kawaref} and for \texttt{UNIT} $50.00$ \cite{El-Kawaref, Hussein-etal-2023-AraBINDER, ElyadataSubmission}. \texttt{CURR} also achieved low performance among all participants ($F_1 \leq66.67$) with exception to \cite{El-Kawaref}, which reported an $F_1=88.89$, despite its low frequency in the data of $25$ occurrences. Our Baseline-II achieved low performance on the three entities mentioned above, but outperformed all submitted models on \texttt{QUANTITY} with an $F_1=75.00$.

For NestedNER, the \texttt{ELYADATA} team~\cite{ElyadataSubmission} ranks in the first position with an $F_1$ score of $93.73$, followed by \texttt{UM6P \& UL} team~\cite{el-mahdaouy-etal-2023-um6p} with a score of $93.09$ and in third place \texttt{AlexU-AIC} with a score of $92.61$. Notably, there are \textit{four} teams that outperform baseline-I with $F_1$ score gap between the baseline and the best model of $2.05$\%. Whereas, the gap between baseline-I and the worst-performing model is about $15.66$\%. The difference in the $F_1$ score among the top four teams is $\sigma=0.57$.

The performance at the entity level for NestedNER is analyzed to explain the challenge for all submitted models. As previously mentioned, the scarcity of some entities in the dataset influences the performance of some entity types in FlatNER. This scarcity influences the results on NestedNER, too. The \texttt{product}, \texttt{quantity}, and \texttt{website} obtained the lowest performance in all models. The highest performance for the \texttt{product} is $66.67$\% which is obtained by ThinkNER team. For the \texttt{quantity}, the $63.16$\% F1-score is obtained by ~\cite{el-mahdaouy-etal-2023-um6p}. For \texttt{website}, the best performance is $69.26$\% F1-score. The \texttt{unit} entity also achieved a low performance among all teams except ~\cite{elkordi-etal-2023-sequence-mrc} which obtained $80$\% F1-score.

The final observation we will highlight is the pattern of scores across the two subtasks, where all scores (micro-F1, precision, and recall) are higher in NestedtNER compared to FlatNER. This was also observed in the baseline \cite{JKG22}. It may seem counter-intuitive, but in fact, FlatNER is harder than NestedtNER. Recall that the Wojood annotation guideline was optimized for nested NER and the flat annotations are simply the top-level tags found in the nested annotations. This conversion from nested to flat annotations caused some tokens to have conflicting tags in the dataset, which breaks the high annotation consistency found in the nested dataset. Another reason for this pattern is the co-occurrence among nested tags. For instance, an entity mention tagged with {\small OCC} is more likely to have nested entity mentions tagged as {\small ORG} or {\small PERS}, rather than entity mentions tagged with {\small PRODUCT}, {\small EVENT} or {\small DATE}.

\subsection{General Description of Submitted Systems}
All the models submitted to the shared task adopt the transfer learning approach, leveraging pre-trained models trained on various data sources. Generally, we observe that the top-performing models addressed the challenge of identifying nested entities of the same type, a limitation described by~\newcite{JKG22}. 

\begin{figure}[h!]  
    \centering       
    \includegraphics[width=0.5\textwidth]{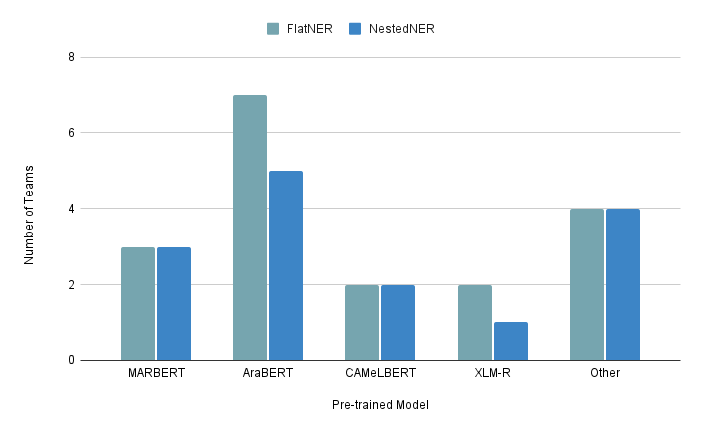}
   \caption{Distribution of pre-trained models across teams.}
\label{fig:pre_trained_model_dist}
\end{figure}

Table \ref{tab:teams_work} summarizes the techniques employed by the participating teams in the WojoodNER-$2023$ shared task. The common theme is the use of pre-trained models by all participants. The choice of models include AraBERT \citep{antoun2020arabert}, MARBERT \cite{abdulmageed2021arbert}, ARBERT \cite{abdulmageed2021arbert}, XLM-R \cite{conneau2019unsupervised}, and CAMelBERT \cite{inoue2021interplay}. AraBART\textsubscript{v2} is the pre-trained language model used the most in the shared task, where it was utilized by seven teams in FlatNER and five teams in NestedNER. MARBERT comes in second place in terms of usage, where six teams used it in both subtasks (Figure \ref{fig:pre_trained_model_dist}).

It was observed in the submissions that compare AraBERT with MARBERT and CAMeLBERT that the AraBERT transformer consistently outperformed the others. This is noteworthy, especially considering that AraBERT is pre-trained solely on MSA data and has a smaller size than both MARBERT and CAMeLBERT.

Other transformer-based pre-trained models were also utilized. For instance, \texttt{Elyadata} fine-tuned BioBERT \cite{lee2020biobert}, but the results were much worse than the baseline, which is expected since BioBERT is trained on English biomedical corpus. In a comparative study, the \texttt{UM6P \& UL}~\cite{el-mahdaouy-etal-2023-um6p} team explored the capabilities of QARiB \cite{abdelali2021pre}, a model pre-trained specifically on Arabic tweets, against ARBERT\textsubscript{v2} \cite{abdulmageed2021arbert}, which is trained on an expansive and diverse Arabic datasets. Their finding shows ARBERT\textsubscript{v2}'s superiority over other models. The rest of this section will discuss the systems submitted by each team in more details.

We start by \texttt{LIPN}~\cite{lipn-at-wojoodner-shared-task} team, who relies on converting the task from sequence labeling to span classification task. Their approach classifies all possible spans within a sequence. For FlatNER, they employ a two-step decoding process: 1) non-entity spans are filtered out, and 2) for the remaining spans, a maximum independent set algorithm is employed to get the optimal set of entity spans. This fusion of algorithmic techniques with machine learning, coupled with the task's reformation, achieved state-of-the-art results for FlatNER and enabled the \texttt{LIPN}~\cite{lipn-at-wojoodner-shared-task} team to secure first place in FlatNER and fourth place in the NestedNER. 

\texttt{UM6P \& UL} \cite{el-mahdaouy-etal-2023-um6p} utilized multi-task learning similar to \cite{JKG22}. The sequence is encoded using a transformer encoder and each entity type has one multi-class classification head to predict the IOB2 tag for each token. The model is trained with multiple objectives including cross-entropy loss, dice loss to handle class imbalance, Tversky loss to balance false positives and false negatives, and focal loss to down-weight easy examples. All four objectives are combined as a weighted sum, the authors refer to the unified loss. Additionally, the authors used variance penalty loss that computes the variance across all task losses. The authors experimented with different loss configurations and pre-trained models, using the unified loss and variance loss with ARBERT\textsubscript{v2} provided the best performance, ranking the team seventh in FlastNER and second in NestedNER. 

\texttt{ELYADATA}~\cite{ElyadataSubmission} team developed the best-performing NestedNER system. They reformulated the task as a denoising problem. DiffusionNER model architecture \cite{shen2023diffusionner} is used with AraBERT, which introduces noise spans to the gold entity boundaries and is trained to reconstruct the entity boundaries. During the inference phase, it picks noisy spans from a standard Gaussian distribution and then produces named entities by leveraging the learned reverse diffusion process. This novel approach enabled the \texttt{ELYADATA}~\cite{ElyadataSubmission} team to get first place and achieve state-of-the-art outcomes in NestedNER. 

\texttt{AlexU-AIC} \cite{elkordi-etal-2023-sequence-mrc} technique relies on machine reading comprehension. In their approach, they formulate a query for each entity type, totaling 21 queries, one for each entity type. Based on the query, the model extracts the answer span from the sequence. Their architecture consists of a transformer encoder followed by two binary classifiers, one classifies if the token is the start of the answer span and another classifies if the token is the end of the answer span. The authors also adopted the stochastic weight averaging technique, in which they average the weights of the four best-performing checkpoints. The team is ranked eighth in FlatNER and third in NestedNER.  

\texttt{AlphaBrains} \cite{ehsanwojoodner2023} developed a multi-task learning technique that is similar to \cite{JKG22}, but it employes BiLSTM encoder instead of a transformer. The input to the BiLSTM is a concatenation of learned word embeddings and ELMo representations. The team is ranked ninth in FlatNER and seventh in NestedNER.

\texttt{El-Kawaref}  \cite{El-Kawaref} proposes StagedNER for FlatNER. In the first stage, the transformer encoder is fine-tuned based IOB2 classification task. In that stage, the authors also used part-of-speech (POS) tagging to improve model performance. The second stage also fine-tunes the transformer encoder on entity type classification task and it takes IOB2 tags as an additional input. During training the authors use the ground truth IOB2 tags and in inference, they use the predicted tags. The team is ranked second in FlatNER. 

\texttt{Alex-U $2023$ NLP} \cite{Hussein-etal-2023-AraBINDER} developed AraBINDER. The approach relies on a contrastive learning objective, where the goal is to maximize the similarity between the entity mention span and its entity type and minimize the similarity with the negative classes. To do that, the authors use a bi-encoder, one for encoding the named entity type and another for encoding the named entity mention. The team is ranked fourth in FlatNER and sixth in NestedNER.


\texttt{Lotus} \cite{lotus2023} proposes a model also inspired by \cite{JKG22}. Their model is based on XLM-R with 21 classification heads, one classifier for each entity type and each classifier is a multi-class that outputs one of the IOB2 tags. The team is ranked tenth in the FlatNER and eighth in the NestedNER.

\section{Conclusion and Future Work}\label{sec:conc}
In this paper, we present the outcomes of WojoodNER-$2023$, the inaugural shared task dedicated to both flat and nested NER challenges in the Arabic language. The results obtained from the participating teams underscore the persistent challenges associated with NER. However, it is promising to observe that various innovative approaches, often harnessing the capabilities of language models, have demonstrated their effectiveness in addressing this complex task. As we move forward, we remain committed to further advancing research in this domain. Our vision includes ongoing efforts to enhance the field of Arabic NER, incorporating the valuable insights gained from WojoodNER-$2023$ and continuing to explore innovative solutions. We plan to extend the Wojood corpus to include more dialects. We plan to include the Syrian Nabra dialects \cite{ANMFTM23} as well as the four dialects in the Lisan \cite{JZHNW23} corpus.

\section*{Acknowledgment}
\label{sec:ack}
We would like to thank Sana Ghanem for helping us with data annotations, and Tymaa Hammouda for her technical support during the organization of the task.

\section{Limitations}\label{sec:limits}
While our aim was to achieve the broadest possible coverage, it is essential to acknowledge that WojoodNER-$2023$ primarily concentrated on MSA data, with only a limited representation of dialects, specifically covering two dialects, Palestinian and Lebanese. 


\normalem
\bibliography{anthology}
\bibliographystyle{acl_natbib}




\end{document}